\documentclass[runningheads]{llncs}
\usepackage[T1]{fontenc}
\usepackage{graphicx}
\usepackage{color,soul}
\usepackage{xcolor}
\usepackage{hyperref}
\usepackage{amsmath}
\usepackage{booktabs} % For better table lines
\usepackage{hyperref}
\usepackage{tabularx} 
\usepackage{siunitx} % Required for alignment
\usepackage{booktabs} % For prettier tables

\usepackage{color}

\begin{document}

	\title{DiRecNetV2: A Transformer-Enhanced Network for Aerial Disaster Recognition}
	\titlerunning{Advancements in Disaster Classification with DiRecNetV2}
	
	\author{Demetris Shianios \orcidID{0009-0005-8266-0727} \and Panayiotis S. Kolios \orcidID{0000-0003-3981-993X} \and Christos Kyrkou*\orcidID{0000-0002-7926-7642}}

	\authorrunning{Demetris Shianios}
	\institute{KIOS Research and Innovation Center of Excellence, University of Cyprus, \\Nicosia, Cyprus\\
		\url{https://www.kios.ucy.ac.cy/} \\
		\email{\{shianios.demetris,kolios.panayiotis,kyrkou.christos\}@ucy.ac.cy}}

	\maketitle              % typeset the header of the contribution

	\begin{abstract}
		The integration of Unmanned Aerial Vehicles (UAVs) with artificial intelligence (AI) models for aerial imagery processing in disaster assessment, necessitates models that demonstrate exceptional accuracy, computational efficiency, and real-time processing capabilities. Traditionally Convolutional Neural Networks (CNNs), demonstrate efficiency in local feature extraction but are limited by their potential for global context interpretation. On the other hand, Vision Transformers (ViTs) show promise for improved global context interpretation through the use of attention mechanisms, although they still remain underinvestigated in UAV-based disaster response applications. Bridging this research gap, we introduce DiRecNetV2, an improved hybrid model that utilizes convolutional and transformer layers. It merges the inductive biases of CNNs for robust feature extraction with the global context understanding of Transformers, maintaining a low computational load ideal for UAV applications. Additionally, we introduce a new, compact multi-label dataset of disasters, to set an initial benchmark for future research, exploring how models trained on single-label data perform in a multi-label test set. The study assesses lightweight CNNs and ViTs on the AIDERSv2 dataset, based on the frames per second (FPS) for efficiency and the weighted F1 scores for classification performance. DiRecNetV2 not only achieves a weighted F1 score of 0.964 on a single-label test set but also demonstrates adaptability, with a score of 0.614 on a complex multi-label test set, while functioning at 176.13 FPS on the Nvidia Orin Jetson device.
		
	\end{abstract}
	\section{Introduction}
	
	The increasing impact of natural disasters such as floods, hurricanes, earthquakes, and wildfires on global regions demonstrates the urgent need for greater awareness and emergency preparedness. Data from Our World in Data \footnote{\href{https://ourworldindata.org/}{https://ourworldindata.org/}} reveal that, on average, natural disasters claim the lives of approximately 45,000 individuals annually, accounting for approximately 0.1\% of deaths worldwide. Regrettably, 2023 encountered several catastrophic events, which comprised the earthquakes in Turkey-Syria that claimed over 33,000 lives and injured thousands more \cite{uwishema2023addressing}, the extensive floods in South Africa and South America \cite{chen2023impacts}, and the devastating wildfires that swept across Western Canada that have burned more than 478,000 hectares of land as of the Canadian Wildland Fire Information System \footnote{\href{https://cwfis.cfs.nrcan.gc.ca/}{https://cwfis.cfs.nrcan.gc.ca/}}. Moreover, the latest global climate science assessment indicates an increasing frequency of simultaneous multiple disasters, presenting complex challenges in disaster management \cite{masson2021climate}. For instance, typhoons can simultaneously cause structural collapses and floods due to their strong winds and rainfall. Likewise, thunderstorms can spark both fires and floods; lightning ignites fires while heavy rains lead to floods. Another complex circumstance is that of earthquakes, which might lead to the collapse of buildings followed by fires in infrastructure. For disaster mitigation and resource allocation to be effective, these multi-label instances must be accurately identified and analyzed.
	
	Unmanned aerial vehicle (UAV) technology such as drones has advanced to the point where it can be an effective tool for minimizing the consequences of natural disasters, particularly in remote regions \cite{9050881}. Through the integration of advanced AI systems and high-resolution cameras, UAVS offer rapid and secure data collection, providing images and videos crucial for mapping disasters, informing authorities promptly, and facilitating emergency response and rescue operations in real-time \cite{9979171}. With the ability to be controlled remotely, they are capable of flying over large, frequently inaccessible areas and turning into intelligent drones that improve situational awareness and operational effectiveness in emergency situations. Moreover, Deep Learning can empower drones to classify disasters, facilitating rapid and precise recognition of impacted zones, evaluation of damage extent, and prioritization of response efforts.

	Deep learning algorithms such as Convolutional Neural Networks (CNNs) have been pivotal in image processing and computer vision, gaining prominence since their inception in 1998 \cite{lecun1998gradient}. More recently, Vision Transformers (ViTs) \cite{dosovitskiy2020image}, have revolutionised image recognition, offering a different approach to interpreting visual data. However, the research community is now exploring the advancements of lighter models in applications like emergency response, where quick processing is essential. These streamlined architectures strike a balance between accuracy and the requirement for quick and effective computation, making them crucial for real-time analysis.

	In this work, we have created a novel hybrid model building on the advantages of CNNs and ViTs, namely the Disaster Recognition Network (\textit{DiRecNetV2}) model. This model represents an advancement from our previous endeavors, in which we introduced DiRecNet, a benchmark dataset (AIDERSv2), and we explored visual explainability techniques like Grad-CAM \cite{shianios2023benchmark}.
	The proposed DiRecNetV2 is designed to address the challenges of accuracy and complexity found in current lightweight models, extending the scope of our ongoing research effort. The proposed model combines the broad contextual strengths of ViTs which help to capture long-range dependencies within an image, with the local inductive biases of CNNs which help to learn hierarchical features. The combination of simple design choices, such as depthwise convolution and a reduced number of heads and encoder blocks in the Vision Transformer, leads to a highly efficient model. This model is specifically designed for the unique needs of UAV-based disaster management. Furthermore, we thoroughly benchmark lightweight CNNs and ViTs on the AIDERSv2 dataset\footnote{link to be provided upon publication}. We evaluated models such as ConvNeXt Tiny \cite{liu2022convnet}, EfficientNet-B0\cite{tan2019efficientnet}, MnasNet \cite{tan2019mnasnet}, MobileNetV2 \cite{sandler2018mobilenetv2}, MobileNetV3 Small \cite{howard2019searching}, ShuffleNetV2 \cite{ma2018shufflenet} and SqueezeNet \cite{iandola2016squeezenet} for the CNNs family, while for ViTs, we examine; Convit Tiny \cite{d2021convit}, GCvit XXtiny \cite{hatamizadeh2023global}, MobileViT \cite{mehta2021mobilevit}, MobileViTV2 \cite{mehta2206separable} and Vit-Tiny \cite{dosovitskiy2020image,steiner2021train}. 
	
	In addition, we propose a multi-label dataset of 300 images, evenly distributed among fire-earthquake, fire-flood, and flood-earthquake disaster combinations. This dataset functions as a test set for deep learning models that are initially trained on more extensive single-label datasets. Our assessments span both single-label and multi-label classification performance, along with efficiency in terms of execution time using the FPS metric (frames processed in a second) over a Jetson Orin device. In our study, DiRecNetV2 has demonstrated outstanding performance on the AIDERSv2 dataset. Additionally, our proposed model has shown robustness in handling complex scenarios, such as identifying multiple disasters in a single image, on a specially constructed multi-label dataset. These results underscore its effectiveness in both single-label and multi-label disaster recognition tasks.
	
	\section{Related Work}
	
	In this section, we delve into related works on disaster classification, including studies exploring the use of lightweight CNNs \cite{9050881}, and the emerging application of Vision Transformers (ViTs) in disaster management scenarios. Several works examine disaster recognition for single classes, such as; earthquakes \cite{ma2019detection,chen2019earthquake,shi2021identifying,ji2020discrimination,ma2020improved,yang2021transferability}, floods \cite{doshi2018satellite,gebrehiwot2019deep,munawar2021uavs,munawar2019after,rahnemoonfar2021floodnet,pally2022application,sarp2020detecting,mao2020train}, and wildfires \cite{shamsoshoara2021aerial,xiong2021fire,jiao2020yolov3,frizzi2021convolutional,hossain2019wildfire,hossain2019wildfire,jadon2019firenet}. Currently, significant research efforts in natural disaster detection have leveraged UAVs, satellites, and social media as primary sources of data.\cite{gadhavi2022transfer,agrawal2021classification,alam2021medic,yuan2023disaster,alam2020deep,bhadra2023mfemanet}.
	
	Multi-label learning addresses the issue of one example being simultaneously linked to multiple labels \cite{zhang2013review}. In the context of a natural disaster, a multi-label problem can be an example of a satellite image that needs to be classified for both flood and wildfire occurrences simultaneously, due to overlapping disaster events in the same geographic region. Another example is a situation where there is a fire and the infrastructure collapses as a result of a bomb explosion. While there are some works regarding multi-label text classification for disaster,  \cite{aipe2018deep,xie2022multi,elangovan2022multi,anggraeni2022deep}, there are few based on computer vision for disaster recognition \cite{singh2022disasternet,cao2023building}. Despite the importance of identifying multiple disasters at once for effective emergency response, there is still a significant research gap concerning the application of Deep Learning models to this challenge.
	
	In recent years, Vision Transformers have been introduced \cite{dosovitskiy2020image}, a transformative method for image classification by adapting the transformer architecture, previously used in natural language processing\cite{vaswani2017attention}. In the field of disaster recognition, only a few studies have been conducted using ViT architecture, focusing on specific areas such as fire detection \cite{khudayberdiev2022fire}, wildfire segmentation \cite{ghali2021wildfire}, flood segmentation \cite{roy2022transformer}, and earthquake magnitude estimation \cite{saad2022real}. There is a dearth of research on the use of Vision Transformers for disaster classification, with a significant gap in understanding their performance in such contexts.
	
	Furthermore, there is a growing need to incorporate AI models into drones and embedded systems for disaster detection/classification purposes \cite{9025436}. For these models to process and analyse streaming and live video footage efficiently, they must be lightweight, ensuring quick inference, low memory usage, and low power consumption. As a result, researchers are investigating more compact architectures of Deep Learning models in disaster management \cite{yuan2022research,yuan2023effc,ge2023lightweight,munsif2022lightweight,yang2022efficient,lee2023watt,mo2023lightweight,saini2023e2alertnet}. Although most of these studies are CNN-based, there has been relatively little investigation into lightweight models for ViTs in the context of disaster recognition tasks.
	
	Overall, many disaster detection approaches focus on single-disaster types, with an ongoing transition towards multi-class detection. However, these models often turn out to be complex for UAV integration, highlighting the necessity for models that align with the limitations of UAV systems. Furthermore, there is a clear gap in models that can handle multi-label disaster scenarios, presenting a vital area for further research and development. Moreover, there is a limited understanding of the effectiveness of Vision Transformers in classifying disasters from aerial images. Our study presents a hybrid CNN-ViT model that integrates lightweight architecture appropriate for embedded systems to precisely fill these gaps. Furthermore, we have developed a multi-label dataset and conducted comprehensive benchmarking to evaluate the performance of lightweight CNNs and Vision Transformers in disaster classification scenarios for both single and multi-label instances.

	\section{Multi-Label Dataset and Hybrid CNN-ViT Model}
	
	The following sections detail the methodological procedure for deriving our two main contributions, the multi-label dataset, and the hybrid CNN-ViT model. 
	
	\begin{figure*}
		\centering
		\includegraphics[width=1\linewidth]{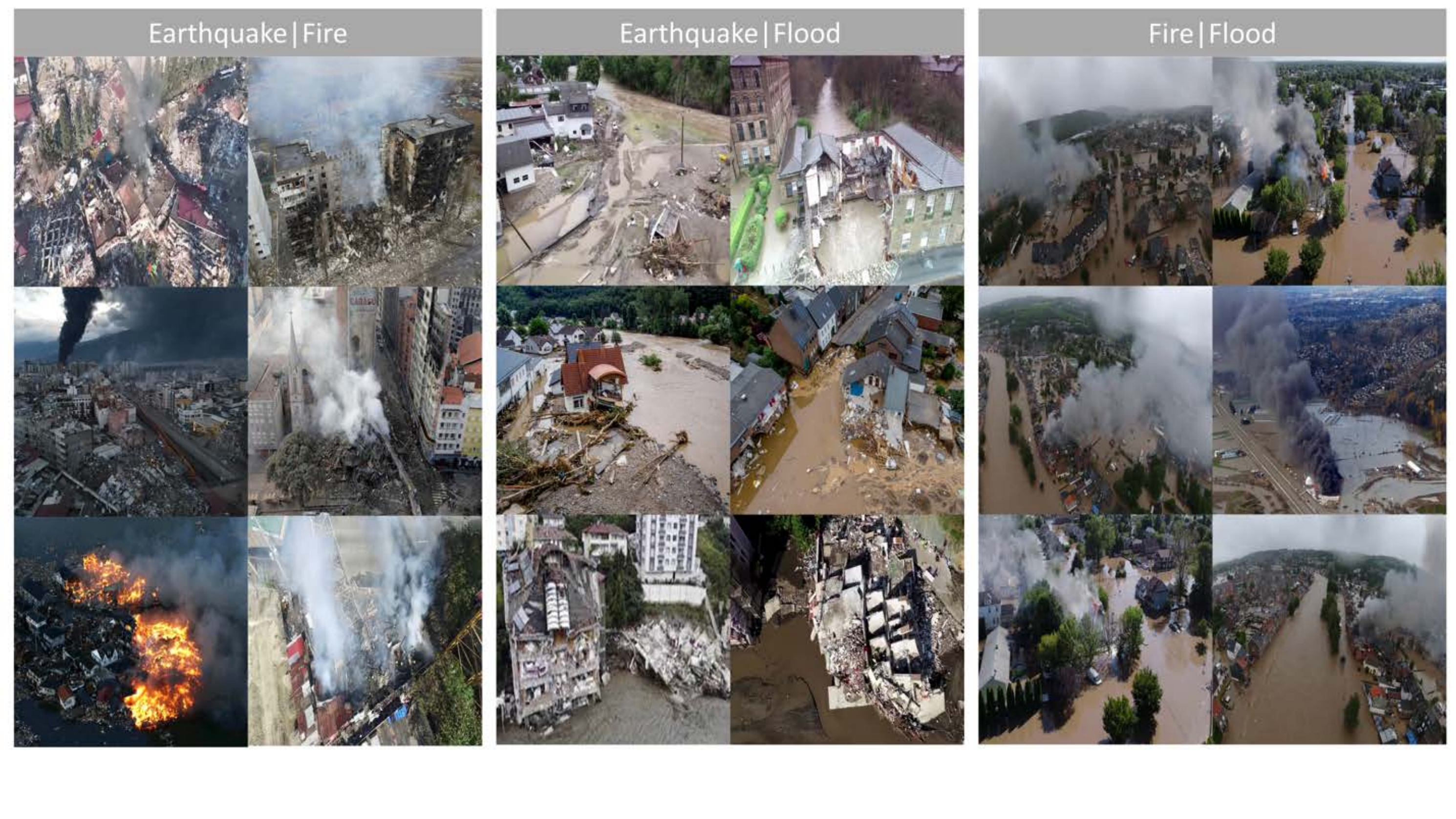}
		\caption{Sample of images in the database depicting various multi-label disaster instances.}
		\label{fig:images_sample}
	\end{figure*}
	
	\subsection{Multi-Disaster Dataset}\label{sec:multi-disaster}

	The development of a multi-label disaster classification system is driven by the essential need to provide accurate, real-time information in response to the occurrence of simultaneous disasters. This is particularly important for facilitating the effective and appropriate distribution of resources and preparedness in multiple disaster situations at the same time. We aim to develop a multi-label dataset that will enable us to explore the capabilities of deep learning models to simultaneously identify multiple disaster events. To curate a dataset for multi-label disaster scenarios, we scoured online platforms such as Google, Bing, and YouTube, using search queries like "flood and fire," "floods and earthquakes/collapsed buildings," and "fires and earthquakes/collapsed buildings." These cases are uncommon, making data collection challenging and time-consuming. Although such scenarios are rare, we collected 300 images, ensuring an equal distribution of 100 images for each disaster combination. 
	
	We carefully examined the chosen images in the ensuing review stage to search for any ambiguities or errors. This thorough validation procedure was essential to maintaining the integrity of the dataset, guaranteeing a low degree of bias, and reducing the possibility of mistakes that would hinder the precision and dependability of our disaster classification model. Two researchers carried out the process of collection and verification. The dataset is currently only used as a test set for models that have already been trained on larger datasets. As the dataset expands, it will become a crucial resource for training and testing models on multi-label instances, thereby improving their accuracy and performance. This work represents a first step, establishing baseline results for models trained on larger single-label datasets and then assessed on our multi-label dataset, providing a foundation for future research endeavors.
	
	\subsection{Single and Multi-Label Classification}
	
	Single-label classification tasks are commonly encountered in numerous Computer Vision applications, with the primary objective being to identify a single result from a range of potential outcomes. This method typically utilizes the softmax activation function (Eq.\ref{eq:soft}) within the final dense layer that outputs the class probabilities. This function guarantees that the total probabilities for all classes sum up to 1, while also ensuring that they are dependent on each other, making it well-suited for single-label classification scenarios.

	\begin{equation}
		\text{softmax}(z)_j = \frac{e^{z_j}}{\sum_{k=1}^{K} e^{z_k}} \label{eq:soft}
	\end{equation}
	
	Where $z$ is the input vector containing logits (raw scores) for each $k$ class. 
	
	In single-label classification training, the commonly used loss function is categorical cross-entropy (Eq.\ref{eq:cce}).
	
	\begin{equation}
		H(y, \hat{y}) = -\sum_{i} y_i \log(\hat{y}_i) \label{eq:cce}
	\end{equation}
	
	The categorical cross-entropy loss function compares the predicted probability $\hat{y}$ distribution with the true distribution $y$ making it suitable for single-label cases where each instance has only one label. \\
	
	On the other hand, when a number of outcomes are possible for a given instance, multi-label classification becomes essential. In that case, the sigmoid activation function (Eq. \ref{eq:sigm})is employed that returns the independent probabilities for each class. Similar to softmax, these probabilities, range between 0 and 1, but are interdependent for each class and might not sum to 1, allowing for simultaneous classification across multiple classes.

	\begin{equation}
		\sigma(z)_j = \frac{1}{1 + e^{-z_j}}\label{eq:sigm}
	\end{equation}
	
	Furthermore, the process of multi-label classification involves converting the labels into one-hot encoded vectors, since they provide a distinct and explicit representation of each label's presence or absence for a particular instance. This encoding format is the key to effectively managing multiple labels per instance, as it enables the binary cross-entropy loss function (Eq. \ref{eq:bce}) to evaluate the prediction accuracy for each individual label independently.

	\begin{equation}
		H(y, \hat{y}) = -\sum_{i} y_i \log(\hat{y}_i) + (1 - y_i) \log(1 - \hat{y}_i) \label{eq:bce}
	\end{equation}

	\subsection{\textbf{Lightweight Deep Learning Models }}

	The development of lightweight deep learning models is increasingly important, particularly for drone-based applications. Given that these models are less resource-intensive, they can operate on devices like drones that have limited processing power. Consequently, intelligence drones, outfitted with lightweight deep learning models, can conduct real-time analysis and make decisions mid-flight, a critical capability for emergency response and disaster monitoring tasks.

	\subsubsection{Convolutional Neural Networks:}

	Convolutional Neural Networks (CNNs), which were first introduced in 1998 \cite{lecun1998gradient} and are well-known for their efficiency in pattern recognition and feature extraction from visual data, have since evolved to become an essential tool in image processing and computer vision applications. However, in an effort to preserve performance while lowering computational load, the trend towards lighter CNN architectures is being driven by the need for efficiency in embedded systems. For quick processing, these simplified models are constructed to maximize the trade-off between accuracy and computational efficiency. They employ strategies such as reduced layer depth, minimized parameters, and compact channel dimensions to simplify their architecture. Techniques like depth-wise separable convolutions are also utilized to cut computational demands by decoupling spatial and channel filtering. Moreover, recent trends in research have moved towards channel attention mechanisms, focusing on salient features to further refine the models' complexity for enhanced performance. In our study, we conducted an analysis of various lightweight CNN models such as;
	ConvNeXt Tiny \cite{liu2022convnet}, EfficientNet-B0\cite{tan2019efficientnet}, MnasNet \cite{tan2019mnasnet}, MobileNetV2 \cite{sandler2018mobilenetv2}, MobileNetV3 Small \cite{howard2019searching}, ShuffleNetV2 \cite{ma2018shufflenet} and SqueezeNet \cite{iandola2016squeezenet}. We decided to include ConvNeXt Tiny in our evaluations even though it might not be generally considered lightweight in comparison to other architectures, however is the lightest model variant within the newly introduced ConvNeXt family.

	\subsubsection{Vision Transformers:}

	Vision Transformers (ViTs) have recently emerged in computer vision applications as potential alternatives to CNNs. Initially, vision transformer models \cite{dosovitskiy2020image}, inspired by their counterparts in language processing \cite{vaswani2017attention}, were computationally heavy with more than 85 million parameters. The research community is now exploring ways to construct lightweight versions of ViTs. The goal of these efforts is to reduce the computing footprint of ViTs while maintaining their advantages, such as their adaptability and global receptive field. By developing lightweight ViTs, researchers hope to bring the advantages of transformer-based models to more resource-limited applications. To reduce the computational demands of self-attention, effective attention techniques like localized or sparse attention are integrated. By factorizing transformer blocks and adding convolutional layers, the architecture is further optimized and a hybrid model that effectively processes spatial information is created. The investigated lightweight ViTs are: Convit Tiny \cite{d2021convit}, GCvit XXtiny \cite{hatamizadeh2023global}, MobileViT-(s,xs,xss) \cite{mehta2021mobilevit}, MobileViTV2-(050,100) \cite{mehta2206separable} and Vit-Tiny \cite{dosovitskiy2020image}.

	\subsection{Transformer Enhanced Convolutional Network Architecture}

	This study integrates a custom CNN with a Vision Transformer (ViT) to present a novel architecture for UAV emergency response. To achieve high performance with minimal computational demands and satisfy the requirements of UAV applications, this hybrid approach combines the effective feature extraction of CNN with the global power capability of ViT. To this end, we end up with the Disaster Recognition Network V2 \textit{DiRecNetV2} model, which is an improvement over our prior work \cite{shianios2023benchmark}, now integrating transformer capabilities.
	
\begin{figure}
		\centering
		\hspace*{-0.05\linewidth} 
		\includegraphics[width=1.1\linewidth]{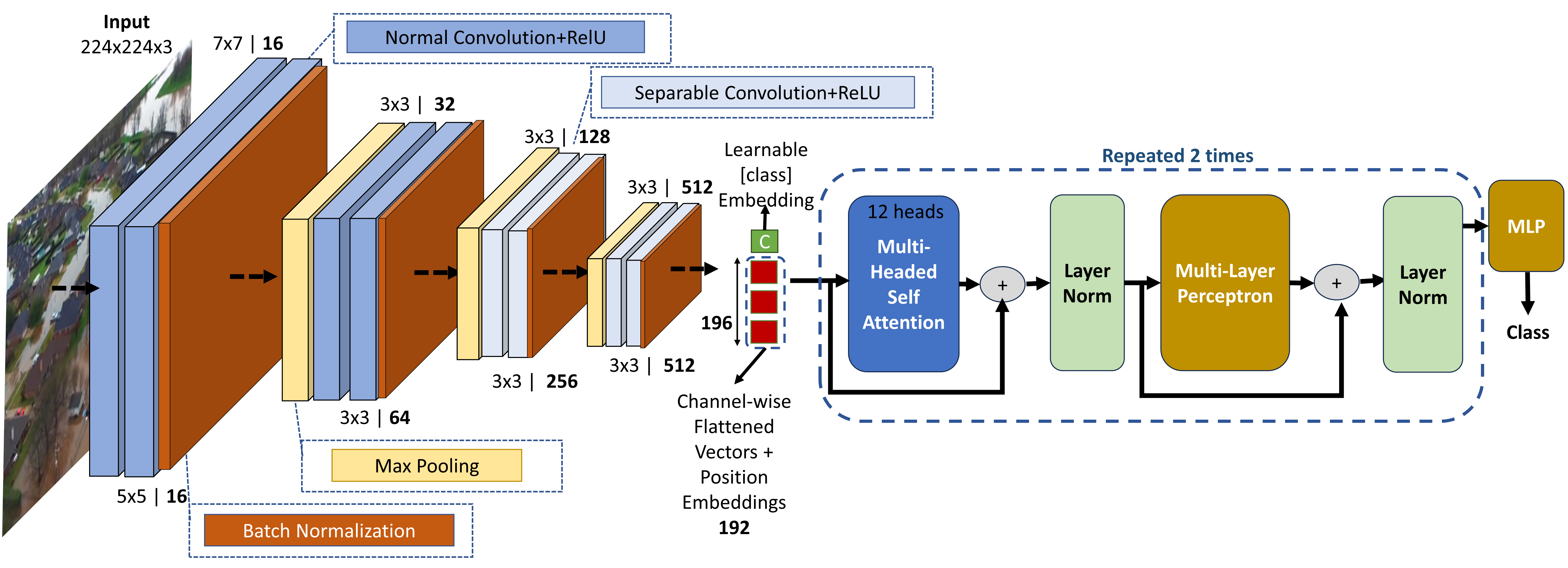}
		\caption{The DiRecNetV2 model architecture, showcasing how features extracted by CNN blocks are fed into the encoder blocks of ViTs of the Vision Transformer, significantly enhancing the model’s capabilities.}
		\label{fig:DiRecNetV2}
	\end{figure}

	The initial model from \textit{DiRecNet} \cite{shianios2023benchmark}, incorporates four key blocks, enabling the learning of hierarchical features while maintaining feature map resolution effectively. In the new version the output of the final block is passed through a transformer encoder block to enhance the model's ability to capture and integrate complex patterns and relationships in the data. This block utilizes the self-attention mechanism inherent in transformers, enabling the model to consider the entire context of the input, leading to a more comprehensive understanding of the features extracted by the DiRecNet backbone. Table \ref{tab:DiRecNetV2} presents a detailed analysis of the model blocks, including the number of parameters, and Fig. \ref{fig:DiRecNetV2} illustrates the model's architecture.
	
	\subsubsection{DiRecNet Feature Extractor:} The model begins with the DiRecNet feature extractor with two consecutive convolutional layers, one with a \(7\times7\) kernel and 16 filters, and the other with a \(5\times5\) kernel and $16$ filters, following modern network trends that apply larger kernels \cite{Liu_2022_CVPR}. Batch normalization and max-pooling of stride $2\times2$ follow. The subsequent block has two \(3\times3\) convolutional layers with $32$ and $64$ filters. After batch normalization, another max-pooling layer is applied. The third block utilizes two separable convolutions with $128$ and $256$ filters, followed by batch normalization and max-pooling. The final block incorporates two identical separable convolutions with $512$ filters, followed by batch normalization and max-pooling. In contrast to the original DiRecNet we add a batch normalization to the last block, a modification absent in the original design. The feature map is subsequently projected to an embedding dimension of $192$, as specified by the transformer hyperparameter. This results in a $192\times14\times14$ feature map, which is then forwarded to the transformer encoder block.
	
	\subsubsection{Transformer Encoder Block: }The feature map is flattened into $196$ element vectors in a depth-wise fashion from $(14 \times 14)$ patches each of 192 elements, creating a denser representation of the image's features. A  classification token 'CLS' is added, resulting in 197 patches, each of 192 dimensions. The patches are passed through a Transformer encoder block with a 12-way multi-head attention mechanism, offering simultaneous focus on various image areas. This approach combines local and global context for comprehensive image analysis. LayerNorm normalization is applied before multi-head attention and before passing through an MLP (Multi-Layer Perceptron) of the same dimensionality of 192. After several experiments, we settled on using two transformer encoder blocks as it resulted in slightly better accuracy. Lastly, the classification head incorporates a LayerNorm as in the standard ViT architecture to normalize the features, and Dropout for regularization. It is followed by a linear layer for class mapping and uses a Softmax in case of single-label classification or a Sigmoid layer for multi-label classification.
	
	\begin{table}[!h]
		\centering
		\footnotesize
		\begin{tabular}{llll}
			\toprule
			\textbf{Layer} & \textbf{Input Shape} & \textbf{Output Shape} & \textbf{Param \#} \\
			\midrule
			\textbf{DiRecNetV2}  & [32, 3, 224, 224] & [32, 4] & 38,016 \\
			\quad \textbf{DiRecNet Feature Extractor}& [32, 3, 224, 224] & [32, 196, 192] & -- \\
			\quad \quad Conv2d  & [32, 3, 224, 224] & [32, 16, 224, 224] & 2,368 \\
			\quad \quad Conv2d  & [32, 16, 224, 224] & [32, 16, 224, 224] & 6,416 \\
			\quad \quad BatchNorm2d & [32, 16, 224, 224] & [32, 16, 224, 224] & 32 \\
			\quad \quad MaxPool2d & [32, 16, 224, 224] & [32, 16, 112, 112] & -- \\
			\quad \quad Conv2d & [32, 16, 112, 112] & [32, 32, 112, 112] & 4,640 \\
			\quad \quad Conv2d & [32, 32, 112, 112] & [32, 64, 112, 112] & 18,496 \\
			\quad \quad BatchNorm2d & [32, 64, 112, 112] & [32, 64, 112, 112] & 128 \\
			\quad \quad MaxPool2d & [32, 64, 112, 112] & [32, 64, 56, 56] & -- \\
			\quad \quad DepthwiseConv2d  & [32, 64, 56, 56] & [32, 64, 56, 56] & 640 \\
			\quad \quad PointwiseConv2d & [32, 64, 56, 56] & [32, 128, 56, 56] & 8,320 \\
			\quad \quad DepthwiseConv2d & [32, 128, 56, 56] & [32, 128, 56, 56] & 1,280 \\
			\quad \quad PointwiseConv2d & [32, 128, 56, 56] & [32, 256, 56, 56] & 33,024 \\
			\quad \quad BatchNorm2d  & [32, 256, 56, 56] & [32, 256, 56, 56] & 512 \\
			\quad \quad MaxPool2d  & [32, 256, 56, 56] & [32, 256, 28, 28] & -- \\
			\quad \quad DepthwiseConv2d & [32, 256, 28, 28] & [32, 256, 28, 28] & 2,560 \\
			\quad \quad PointwiseConv2) & [32, 256, 28, 28] & [32, 512, 28, 28] & 131,584 \\
			\quad \quad DepthwiseConv2d & [32, 512, 28, 28] & [32, 512, 28, 28] & 5,120 \\
			\quad \quad PointwiseConv2d & [32, 512, 28, 28] & [32, 192, 28, 28] & 98,496 \\
			\quad \quad BatchNorm2d & [32, 192, 28, 28] & [32, 192, 28, 28] & 384 \\
			\quad \quad MaxPool2d  & [32, 192, 28, 28] & [32, 192, 14, 14] & -- \\
			\quad \quad Flatten  & [32, 192, 14, 14] & [32, 197, 192] & -- \\
			\quad Dropout  & [32, 197, 192] & [32, 197, 192] & -- \\
			\quad \textbf{Transformer Encoder Blocks} & [32, 197, 192] & [32, 197, 192] & -- \\
			\quad \quad TransformerEncoderBlock (1) & [32, 197, 192] & [32, 197, 192] &  -- \\
			\quad \quad \quad MultiheadSelfAttentionBlock & [32, 197, 192] & [32, 197, 192] & 148,608 \\
			\quad \quad \quad MLPBlock & [32, 197, 192] & [32, 197, 192] & 74,496 \\
			\quad \quad TransformerEncoderBlock (2) & [32, 197, 192] & [32, 197, 192] & -- \\
			\quad \quad \quad MultiheadSelfAttentionBlock  & [32, 197, 192] & [32, 197, 192] & 148,608 \\
			\quad \quad \quad MLPBlock & [32, 197, 192] & [32, 197, 192] & 74,496 \\
			\quad \textbf{Classifier Head} & [32, 197, 192] & [32, 4] & -- \\
			\quad \quad LayerNorm  & [32, 192] & [32, 192] & 384 \\
			\quad \quad Dropout  & [32, 192] & [32, 192] & -- \\
			\quad \quad Linear  & [32, 192] & [32, 4] & 772 \\
			\quad \quad Softmax/Sigmoid & [32, 4] & [32, 4] & -- \\
			\bottomrule
		\end{tabular}
		\caption{The DiRecNetV2 model’s structured layout, showcases the evolution of feature spaces across various blocks, complete with input and output dimensions utilizing a batch size of 32. In addition, we can explore, the number of parameters at each stage, culminating in a streamlined model architecture with a total of 799,380 parameters.}
		\label{tab:DiRecNetV2}
	\end{table}

	\subsection{Training Process}
	
	\subsubsection{Datasets:}

	In our research, we initially divided the data, sourced from the AIDERSv2 dataset as described in \cite{shianios2023benchmark}, into training, validation, and test sets, with proportions of 80\%, 10\%, and 10\% respectively. The distribution of these sets is detailed in Table \ref{tab:data_proportion}. 
	For the multi-label evaluation, it is important to highlight that the training was performed on the same dataset as used for the single-label tasks, with the only difference being in the final activation function. However, the test set was the Multi-label dataset discussed in Section \ref{sec:multi-disaster}. We trained the models on data with single disasters for the purpose to assess their performance on new images with more than one disaster. Future work could involve fine-tuning on the multi-label dataset; once we acquire a more extensive multi-disaster dataset. We can split it into training, validation, and test sets to investigate how model performance varies when fine-tuned on a multi-label dataset during training.

	\begin{table}[!h]
		\centering
		\scalebox{0.95}{%
			\begin{tabular}[c]{ c | c | c | c | c | c}
				\hline
				\centering
				& \textbf{Earthquakes} & \textbf{Floods} & \textbf{Wildfire/Fire} & \textbf{Normal} & \textbf{Total} \\[1ex]
				\hline
				\textbf{Train} & 1927 & 4063 & 3509 & 3900 & 13399 \\
				\textbf{Validation} & 239 & 505 & 439 & 487 & 1670 \\
				\textbf{Test} & 239 & 502 & 436 & 477 & 1654 \\
				\hline
				\textbf{Total} & 2405 & 5070 & 4384 & 4864 & \textbf{16723} \\
				\hline
			\end{tabular}
		}
		\caption{Proportion of images in each class within the train, validation, and test set.}
		\label{tab:data_proportion}
	\end{table}

	\subsubsection{Data Pre-Processing:}
	
	The images were scaled to $224\times224\times3$ and standardized for DiRecNetV2; therefore, to change the distribution to have a mean of zero and a standard deviation of one. Random augmentations were applied to expand the diversity of the dataset and combat overfitting. Specifically, we applied rotation, zoom, horizontal shift, vertical shift, horizontal flip, and shear.

	\subsubsection{Transfer Learning:}

	For training the different baseline networks we adopted a transfer learning approach for both lightweight Convolutional Neural Networks (CNNs) and Vision Transformers (ViTs). This method significantly accelerated the training process by leveraging the knowledge embedded in the weights obtained from training these models on the extensive ImageNet dataset \cite{krizhevsky2017imagenet}. The use of transfer learning offers a substantial advantage as it uses pretrained models with a rich understanding of diverse features from larger dataset, thereby enhancing learning efficiency and performance on diverse tasks. Our sole modification involved adapting the classifier head for each pretraiend lightweight model. Specifically, to mitigate overfitting, we incorporated a dropout layer with a rate of 0.5, followed by a dense layer tailored to handle the number of classes in our case to four.
	
	\subsubsection{Training Regime:}
	
	For our custom-designed model that was not pretrained on ImageNet, we conducted training of 300 epochs from scratch to refine its learning capabilities. In contrast, for the pretrained models, we limited the training to 40 epochs, considering their prior knowledge acquired from the larger dataset. This training strategy was uniformly applied across both single-label and multi-label classification experiments. In all cases, we utilized the Adam optimizer with a learning rate of  $1e-4$ and batch size was set to 32, while for the model selection we chose the iteration of the model where the validation accuracy was at its highest akin to early stopping.
	
	\subsubsection{Configurations:}
	The experiments were carried out on the Linux operating system using the Tesla V100 Graphics Processor Unit, with 64GB RAM and CUDA version 10.2. We use PyTorch \footnote{\href{https://pytorch.org/}{https://pytorch.org/}} 1.12.1 as the deep learning framework along with Python \footnote{\href{http://www.python.org}{http://www.python.org}} version 3.8.0. 
	Additionally, to evaluate the models' frames per second (FPS), we deploy them on NVIDIA's jetson Orin device.

	\section{Evaluation and Results}

	\subsection{\textbf{Evaluation Metrics}}
	
	To evaluate the performance of the models, we investigated two key performance indicators since both accuracy and speed are crucial to detect natural disasters in real time. These are the weighted F1 score (Eq.\ref{eq:wf1}) and frames per second (FPS). We use Weighted F1-Score for both single-label and multi-label evaluation, because it accounts for the relative importance of each label by weighting the F1 Score of each label according to its prevalence in the dataset. The FPS values, acquired from tests conducted on the NVIDIA Jetson Orin device.

	\begin{equation}
		\text{Weighted F1 Score} = \sum_{i=1}^{N} w_i \times \text{F1 Score}_i \label{eq:wf1}
	\end{equation}
	\text{Where,}
	\begin{equation*}
		w_i = \frac{\text{No. of samples in class } i}{\text{Total number of samples}}
	\end{equation*}

	We use the scoring scheme of our previous research \cite{shianios2023benchmark} to determine the best trade-off between speed and accuracy in model performance, as defined in the Eq. \ref{eq:eq_score}. We can determine the model that provides the best fit for a particular application scenario by varying the parameter $\lambda$. In the current study, we have given the lambda a value of $0.3$ when we prioritize speed and $0.7$ when we prioritize accuracy. Moreover, for a broader assessment, we compare the models against a revised scoring equation from \cite{ignatov2021fast}, as presented in Eq. \ref{eq:eq_score2}. To match the scale of the evaluation metrics, we have set the normalization constant $C$ at $1e27$ in this instance.

	\begin{align}
		\text{Score1} &= \lambda\times F1_{\text{norm}} + (1-\lambda)\times FPS_{\text{norm}} \label{eq:eq_score} \\
		\text{Score2} &= \frac{2^{\text{F1}} \times \text{FPS}}{C} 
		\label{eq:eq_score2}
	\end{align}
	
	However, prior to this, we use the formula in Eq. \ref{eq:eq_norm} to normalize the values of FPS and Weighted F1 across all models since their ranges differ. Values in $x$ are squeezed into the range $[a,b]$, where $a$ was set to 0.1 and $b$ at 1, making the variables comparable to one another.

	\begin{equation}
		x_{norm} = (b-a)\frac{x - min(x)}{max(x) - min(x)} + a 
		\label{eq:eq_norm}
	\end{equation}

	\begin{table}[!h]
		\centering
		\footnotesize
		
		\begin{tabular}{|p{4.4cm}|c|c|c|c|}
			\hline
			\textbf{Model Name} & \textbf{GFLOPs} & \textbf{No. of} & \textbf{Model Size} & \textbf{FPS} \\
			&                 & \textbf{Params (M)} & \textbf{(MB)}        & (1/s) \\
			\hline
			Convit Tiny (2021) \cite{d2021convit} & 1.08 & 5.52 & 22.07 & 54.34 \\
			GCVit XXtiny (2023) \cite{hatamizadeh2023global} & 1.94 & 11.48 & 45.93 & 31.06\\
			MobileViT s (2021) \cite{mehta2021mobilevit} & 1.42 & 4.94 & 19.76 & 30.71\\
			MobileViT xs  (2021)\cite{mehta2021mobilevit} & 0.71 & 1.93 & 7.74 & 26.07\\
			MobileViT xxs (2021) \cite{mehta2021mobilevit} & 0.26 & 0.95 & 3.81 & 33.05 \\
			MobileVitV2 050 (2022) \cite{mehta2206separable} & 0.36 & 1.11 & 4.46 & 33.42 \\
			MobileVitV2 0100 (2021) \cite{mehta2206separable} & 1.41 & 4.39 & 17.56 & 37.65\\
			Vit Tiny  \cite{dosovitskiy2020image,steiner2021train} 
			& 1.08 & 5.53 & 22.1 & 74.88
			\\
			ConvNext Tiny (2022)\cite{liu2022convnet}
			& 4.46 & 27.82 & 111.29 & 89.09\\
			EfficientNet-B0 (2019) \cite{tan2019efficientnet} & 0.41 & 4.01 & 16.05 & 55.74\\
			MnasNet (2019) \cite{tan2019mnasnet} & 0.34 & 3.11 & 12.43 & 90.98 \\
			MobileNetV2 (2018) \cite{sandler2018mobilenetv2} & 0.33 & 2.23 & 8.92 &  87.23 \\
			MobileNetV3 Small (2019) \cite{howard2019searching} & 0.06 & 0.93 & 3.72 & 89.72\\
			ShuffleNetV2 (2017) \cite{ma2018shufflenet} & 0.15 & 1.26 & 5.03 & 75.73\\
			SqueezeNet (2016) \cite{iandola2016squeezenet} & 0.26 & 0.73 & 2.90 & 
			183.08 \\
			DiRecNetV2 (Proposed) & 1.09 & 0.80 & 3.20  & 176.13\\
			\hline
		\end{tabular}
		\caption{Comparison of lightweight Vision Transformers (ViTs) and lightweight CNN models, focusing on their computational complexity by giga Flops (floating-point operations per second), the number of parameters, and model size. Each model has been customized with a classifier head, including a dropout rate (0.5) between the feature extractor and the fully connected layer. This customization is aligned with the target class count, which in this case is four.}
		\label{tab:vits_cnns_comparison}
	\end{table}

	\subsection{Computational Efficiency Evaluation}

	In our analysis comparing deep learning architectures for real-time disaster recognition, the introduced DiRecNetV2 model stands out as it has the smallest number of parameters at just 0.8 million. With the exception of ConvNeXt, which has 27.82 million parameters, the other examined models have parameters under 12 million. In terms of computational demand, measured in giga FLOPs, all models are below 2, except for ConvNeXt which has 4.46. When considering the size of the models, we observe a wider distribution; DiRecNetV2, at 3.20 MB, has the second-smallest footprint, closely followed by SqueezeNet's 2.90 MB. DiRecNetV2's low parameter count makes it particularly well-suited for embedded devices where resource efficiency is crucial. 
	
	Moreover, the examination of model's execution reveals that DiRecNetV2's processing speed is second with 176.13 FPS only to SqueezeNet, which leads with an FPS of 183.08 but at a higher parameter count and larger size. This approach is especially advantageous for high-speed drones that are tasked with wide-area surveillance missions, where there is a requirement to analyze an extensive number of frames. In such scenarios, the ability to rapidly process and interpret vast amounts of visual data is crucial. A detail analysis of the computational efficiency of the models is presented in Table \ref{tab:vits_cnns_comparison}.

	\subsection{Disaster Classification Evaluation}

	\begin{table}[htbp]
		\centering
		\begin{tabular}{|l|c|l|l|l|l|}
			\hline
			\textbf{Model Name} & \multicolumn{1}{|c|}{\textbf{Weighted}} & \multicolumn{3}{c|}{\textbf{Score1}} & \multicolumn{1}{c|}{\textbf{Score 2}} \\
			\cline{3-5}
			& \textbf{F1} & \multicolumn{1}{|c|}{\textbf{Balance}} & \multicolumn{1}{c|}{\textbf{Prioritize WF1}} & \multicolumn{1}{c|}{\textbf{Prioritize FPS}} & \multicolumn{1}{c|}{} \\
			\hline
			ConvNeXt Tiny          & 0.940 & 0.646 & 0.720 & 0.572 & 1764.609 \\
			EfficientNet-B0             & 0.862 & 0.281 & 0.285 & 0.277 &    4.988 \\
			MnasNet                 & 0.897 & 0.502 & 0.513 & 0.490 &   89.600 \\
			MobileNetV2             & 0.893 & 0.479 & 0.490 & 0.467 &   67.401 \\
			MobileNetV3 Small      & 0.868 & 0.398 & 0.372 & 0.425 &   12.003 \\
			ShuffleNetV2            & 0.852 & 0.304 & 0.272 & 0.336 &    3.436 \\
			SqueezeNet              & 0.845 & 0.586 & 0.421 & 0.752 &    4.974 \\
			Convit Tiny           & 0.871 & 0.307 & 0.324 & 0.289 &    8.827 \\
			GCVit XXtiny          & 0.932 & 0.452 & 0.582 & 0.323 &  355.801 \\
			MobileVitV2 050        & 0.834 & 0.113 & 0.108 & 0.119 &    0.403 \\
			MobileVitV2 100       & 0.875 & 0.240 & 0.297 & 0.184 &    5.705 \\
			MobileViT s          & 0.855 & 0.190 & 0.210 & 0.170 &    1.759 \\
			MobileViT xs           & 0.837 & 0.131 & 0.126 & 0.135 &    0.532 \\
			MobileViT xxs        & 0.859 & 0.219 & 0.241 & 0.198 &    2.775 \\
			Vit Tiny & 0.873 & 0.375 & 0.373 & 0.377 &   14.666 \\
			DiRecNetV2              & \textbf{0.964} & \textbf{0.980} & \textbf{0.988} & \textbf{0.972} & \textbf{18982.892} \\
			\hline
		\end{tabular}
		\caption{Evaluating model efficiency in contexts where accuracy is crucial against
			situations demanding high frames per second (FPS) highlights different operational
			focuses on emergency management and live surveillance.}
		\label{tab:model_performance}
	\end{table}

	\begin{figure}
		\centering
		\hspace*{-0.05\linewidth} 
		\includegraphics[width=1.1\linewidth]{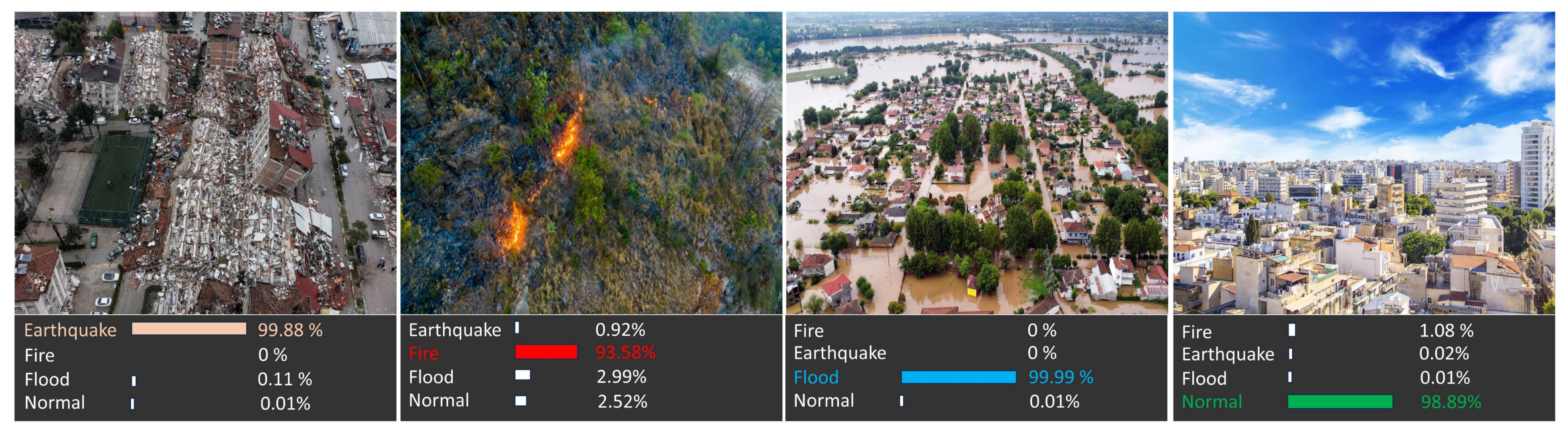}
		\caption{The examples demonstrate DiRecNetV2's proficiency in identifying diverse disaster situations. Using a subset of four test set images, these images show the model's robust classification accuracy for earthquakes, fires, floods, and normal cases.}
		\label{fig:preds}
	\end{figure}

	\subsubsection{Single-Label Classification Performance: } 
	
	In our comparative analysis of lightweight CNNs and ViTs we observed that there was a range in performance based on the weighted F1 score. Scores for the lightweight CNNs range, from 0.845 of the SqueezeNet to the highest performance at 0.940 of ConvNeXt Tiny. Lightweight ViTs exhibited a performance spectrum, with scores spanning from 0.837 for MobileViT xs to 0.932 for GCViT XXtiny. This analysis suggests that lightweight Vision Transformers have the potential to compete with traditional CNNs in the task of disaster identification from aerial images. Notably, within both families, the only two models that achieved a score above 0.90 were the more complex ones (ConvNeXt, GCViT XXtiny), characterized by a larger number of parameters and greater model size. This pattern suggests that although efficient lightweight models can be obtained, there is still an association between higher performance and greater complexity.

	Conversely, with its simplified architecture and fewer parameters, the DiRecNetV2 model achieved remarkable results, outperforming all other models with a notable weighted F1-Score of 0.964. This superior performance was further confirmed through the evaluations using the scoring formulas tailored to assess both speed and accuracy. The DiRecNetV2 model excelled in three critical scenarios: where accuracy and speed are equally prioritized, where speed is paramount, and where accuracy is of most importance as shown by the results in Table \ref{tab:model_performance}. When equal emphasis is placed on speed and performance, the proposed model attains a score of 0.980, followed by ConvNeXt, which scores 0.646. In scenarios where the focus is on maximizing the weighted F1 score, the proposed model again leads with a score of 0.988, followed by ConvNeXt with a score of 0.720. Moreover, when the priority shifts to frames per second (FPS), our model maintains the highest performance with a score of 0.972, with SqueezeNet trailing behind at 0.752. Furthermore, when evaluating model performance using the Score2 (Eq.\ref{eq:eq_score2}) DiRecNetV2 again surpasses competing models. Therefore, the proposed model manages to bridge the gap between high accuracy and low efficiency, delivering exceptional classification performance and execution speed. Examples of visual predictions are shown in Figure \ref{fig:preds}.
	
	\subsubsection{Multi-Label Classification Performance: }

	During the multi-label evaluation, we applied a threshold of 0.5 to each class's probability output, post-sigmoid activation, meaning a class was assigned if its probability was higher than this threshold. The evaluation of the different models in the multi-label classification task shows a common trend where precision is consistently high for all models over the three disaster classes. This suggests that the models perform well in terms of specificity; where the instances they label positively for a disaster type are correct, demonstrating a low false positive rate. Recall scores, on the other hand, are less consistent and typically lower in contrast, indicating that the models may not be as good at finding all pertinent examples of each class. This suggests that even though the models accurately predict outcomes, a significant number of cases or "true positives", the instances correctly classified as belonging to a specific class,  are being neglected. The reason for the low recall rates observed could be that the models were not optimized for datasets with simultaneous disaster events, thus models are less likely to identify individual disasters and ignore situations where features from multiple disasters coexist in one image. In general, the results suggest that when the models were fine-tuned to transition from recognizing single disaster events to detecting dual disaster scenarios, their performance diminished due to the increased complexity and overlap of disaster characteristics.

\begin{figure}
		\centering
		\includegraphics[width=1\linewidth]{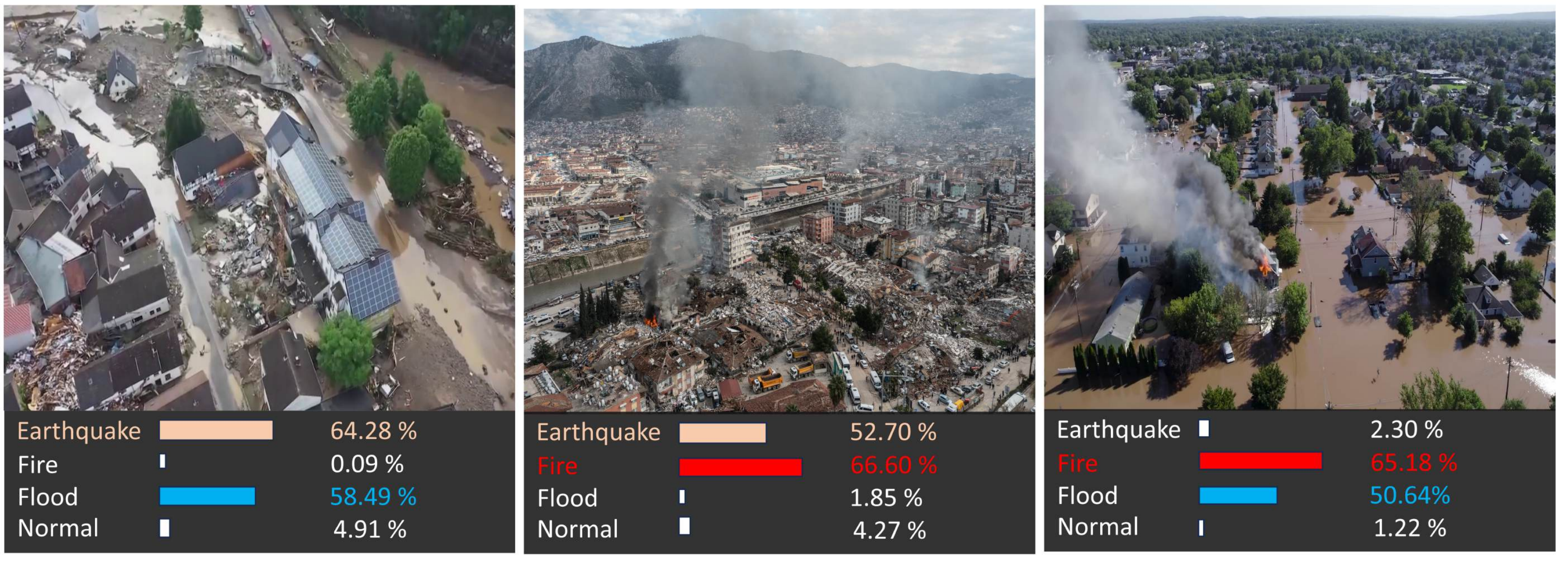}
		\caption{Examples of images from the multi-label dataset showcase the predictions of DiRecNetV2, trained for multi-label scenarios. The predictions illustrate the model's accurate identification of dual instances, with probabilities exceeding 50\% for two classes within the same image, underscoring its proficiency in handling complex multi-label classifications.}
		\label{fig:preds_ml}
	\end{figure}

	Comparison of the models reveals that the DiRecNetV2 stands out with stronger performance across the weighted F1-score which is the harmonic mean of precision and recall. Specifically, it achieves the highest average Weighted F1 metric of 0.614 indicating that DiRecNetV2 maintains a balance between precision and recall compared to other models, making it the most reliable choice among those listed for multi-label disaster classification.
	An important observation is the lower recall value of 0.230 for fires compared to earthquakes and floods in the multi-label model is probably due to how the model was trained. During training, the model had been exposed primarily to single-label images where the entire image was associated with one disaster type, thus enabling it to learn distinctive features relevant to that particular disaster. Consequently, when encountering images with multiple disaster types during testing, such as an image featuring both fires and floods, the model may struggle to identify the less prominent fire-related features, leading to a reduced recall rate for the fire category. DiRecNetV2's performance highlights the possibilities of hybrid models that combine transformer and convolutional-based architectures, not only for single-label classification tasks but also for multi-label ones. It would be interesting to discover how these models' performance changes in subsequent work when they are fine-tuned on a multi-label dataset. Table \ref{tab:model_performance_ml} summarize the results of multi-label classification, while visual examples are presented in Figure \ref{fig:preds_ml}.

	\begin{table}[!h]
		\centering
		\resizebox{\textwidth}{!}{%
			\begin{tabular}{|l|c|c|c|c|c|c|c|c|c|c|}
				\hline
				\textbf{ModelName} & \multicolumn{3}{|c|}{\textbf{Earthquakes}} & \multicolumn{3}{c|}{\textbf{Fires}} & \multicolumn{3}{c|}{\textbf{Flood}} & \textbf{Average} \\
				\cline{2-11}
				& {Precision} & \multicolumn{1}{c}{Recall} & \multicolumn{1}{c|}{Weighted F1} & {Precision} & \multicolumn{1}{c}{Recall} & \multicolumn{1}{c|}{Weighted F1} & {Precision} & \multicolumn{1}{c}{Recall} & \multicolumn{1}{c|}{Weighted F1} & {Weighted F1} \\
				\hline
				ConvNext Tiny & 1.000 & 0.485 & 0.653 & 1.000 & 0.695 & 0.820 & 1.000 & 0.180 & 0.305 & 0.593 \\
				EfficientNet-B0  & 1.000 & 0.590 & 0.742 & 1.000 & 0.100 & 0.182 & 1.000 & 0.355 & 0.524 & 0.483 \\
				MnasNet & 0.982 & 0.560 & 0.713 & 1.000 & 0.195 & 0.326 & 1.000 & 0.260 & 0.413 & 0.484 \\
				MobileNetV2  & 1.000 & 0.440 & 0.611 & 1.000 & 0.270 & 0.425 & 1.000 & 0.195 & 0.326 & 0.454 \\
				MobileNetV3 & 1.000 & 0.585 & 0.738 & 1.000 & 0.285 & 0.444 & 1.000 & 0.285 & 0.444 & 0.542 \\
				ShuffleNetV2 & 1.000 & 0.285 & 0.444 & 1.000 & 0.050 & 0.095 & 1.000 & 0.110 & 0.198 & 0.246 \\
				SqueezeNet & 1.000 & 0.550 & 0.710 & 1.000 & 0.165 & 0.283 & 1.000 & 0.290 & 0.450 & 0.481 \\
				Convit Tiny & 1.000 & 0.350 & 0.519 & 1.000 & 0.665 & \textbf{0.799} & 1.000 & 0.145 & 0.253 & 0.524 \\
				GCVit XXtiny & 1.000 & 0.325 & 0.591 & 1.000 & 0.650 & 0.788 & 0.960 & 0.120 & 0.213 & 0.497 \\
				MobileViT s & 0.920 & 0.630 & 0.748 & 0.984 & 0.305 & 0.466 & 0.977 & 0.430 & 0.597 & 0.604 \\
				MobileViT xs  & 0.780 & 0.640 & 0.703 & 1.000 & 0.285 & 0.444 & 0.925 & 0.430 & 0.587 & 0.578 \\
				MobileViT xxs & 0.862 & 0.685 & 0.763 & 1.000 & 0.185 & 0.312 & 0.940 & 0.625 & \textbf{0.751} & 0.609 \\
				MobileVitV2 050 & 0.850 & 0.735 & 0.788 & 1.000 & 0.305 & 0.467 & 1.000 & 0.270 & 0.425 & 0.560 \\
				MobileVitV2 0100 & 0.916 & 0.655 & 0.764 & 1.000 & 0.255 & 0.406 & 0.976 & 0.200 & 0.332 & 0.501 \\
				Vit Tiny & 1.000 & 0.460 & 0.630 & 1.000 & 0.640 & 0.780 & 0.872 & 0.205 & 0.332 & 0.581 \\
				DiRecNetV2\ & 1.000 & 0.605 & \textbf{0.754} & 1.000 & 0.280 & 0.438 & 0.926 & 0.500 & 0.649 & \textbf{0.614} \\
				\bottomrule
			\end{tabular}
		}
		\caption{Model performance comparison for multi-label classification.}
		\label{tab:model_performance_ml}
	\end{table}
	
	\subsection{Accuracy-Efficiency Spectrum Gap}
	
	The experimental results highlight that architectures of greater complexity achieve greater classification performance, whereas architectures of lesser complexity exhibit higher efficiency and faster execution for both the CNNs and ViTs deep learning frameworks. Our findings serve as a benchmark and demonstrate the gap researchers need to address to design models achieving high classification performance while maintaining FPS within the same order of magnitude. With this objective in mind, our proposed DiRecNetV2 model is designed to bridge this gap.

	DiRecNetV2's architecture combines the strength capabilities of standard convolutions, the efficiency of depth-wise separable convolutions, and the global contextual understanding afforded by the attention mechanism. Unlike the typical vision transformer approach that splits the image into patches, our methodology employs CNN's feature extraction vectors, which are then fed into the encoder block of the vision transformer. The results indicate that such a hybrid model is not only suitable but potentially superior for the task of disaster recognition from aerial images, where the integration of both local and global features is paramount. The DiRecNetV2 performance demonstrates the effective combination of high classification accuracy, rapid processing (high fps), and low complexity, setting a new benchmark for efficient and accurate image classification tasks. Moreover, architectures like DiRecNetV2 show promise for multi-label scenarios, where images may contain multiple events, as our model also outperforms competitors in multi-label test datasets.

	\section{Conclusion and Future Work}

	Advances in machine learning and computer vision anticipate a new era of technological empowerment for humanitarian relief by providing tools that improve the efficacy and speed of life-saving steps in times of disaster. At the same time, the integration of unmanned aerial vehicles (UAVs), like drones, with cutting-edge deep learning algorithms marks a significant step forward in enhancing disaster relief efforts. In this research, we propose a hybrid model that combines the benefits of Convolutional Neural Networks (CNNs) and Vision Transformers (ViTs) to develop an accurate, resource-efficient framework suitable for UAV-based disaster detection applications. We introduce the DiRecNetV2, that demonstrates remarkable results by fusing the global contextual awareness of ViTs with the feature extraction power of CNNs. This model not only achieves the highest accuracy among the evaluated models but also ranks second in terms of processing speed. Implementing evaluation criteria that take into account both processing speed and accuracy using two distinct scoring systems, we proved that our model is particularly suitable for embedded systems. This result highlights the effectiveness and efficiency of such hybrid architectures in real-world applications, emphasizing their potential for future advances in a variety of computer vision tasks. By integrating CNNs and ViTs into a lightweight framework, the hybrid model offers promising solutions for resource-constrained environments, such as edge computing and mobile applications.

	Additionally, we present benchmark results for both the AIDERSv2 dataset and our newly introduced multi-label dataset, which comprises 300 images featuring overlapping disasters. These benchmarks encompass performance evaluations for lightweight CNNs and ViTs. To illustrate how these models perform when trained on single-label datasets and then assessed on the proposed multi-label dataset, we provide baseline performances for these models. An important aspect of our study is the provision of benchmark results for ViTs, which remains underexplored in the field of multi-disaster recognition within existing literature.  The promising performance of DiRecNetV2 in multi-label tasks underscores the effectiveness of this hybrid approach in handling such scenarios, suggesting its potential applicability in various domains such as environmental monitoring, and industrial safety.  For instance, in agricultural monitoring, multi-label identification could help detect various crop diseases and nutrient deficiencies from drone or satellite imagery. Similarly, in urban planning, the model's ability to identify multiple urban features like buildings, roads, and green spaces from satellite or aerial imagery streamlines city development and infrastructure management processes.

	In future studies, we intend to grow the size of the multi-label dataset and fine-tune the examined lightweight model, evaluating the effects of training on this enriched multi-label dataset as opposed to a single-label one on performance. We also intend to explore the explainability side of these algorithms, investigating how CNNs and ViTs differ in feature recognition and comprehending how hybrid models discriminate between different image features in classification tasks. This study will primarily focus on the attention mechanisms in ViTs and hybrid models, providing insights into how they prioritize distinct image regions or features throughout the decision-making process. We expect that our proposed DiRecNetV2 model on the AIDERSv2 and multi-label datasets, combined with the benchmark results of lightweight CNNs and ViTs, will provide a solid foundation for future research. This endeavor aims to encourage the advancement of novel approaches for the application of disaster response, ultimately contributing to the communities impacted by such situations.

	\section*{Acknowledgements}
	This work is supported by the European Union Civil Protection Call for proposals UCPM-2022-KN grant agreement No 101101704 (COLLARIS Network). The work is partially supported by the European Union’s Horizon 2020 research and innovation program under grant agreement No 739551 (KIOS CoE - TEAMING) and from the Republic of Cyprus through the Deputy Ministry of Research, Innovation and Digital Policy.\\
	Christos Kyrkou would like to acknowledge the support of NVIDIA with the donation of GPU platform.
	
	\section*{Declarations}
	
	\begin{itemize}
		\item \textit{Funding:} See Acknowedgements Section
		\item \textit{Conflict of interest/Competing interests:} Not Applicable
		\item \textit{Ethics approval and consent to participate:} Not Applicable
		\item \textit{Consent for publication:} Not Applicable
		\item \textit{Data availability:} Immediately with paper publication
		\item \textit{Materials availability:} Not Applicable
		\item \textit{Code availability:} Immediately with paper publication
		\item \textit{Author contribution:} Demetris Shianios made significant contributions to this project through implementation and writing. Christos Kyrkou provided guidance, mentoring, and idea formulation. Panayiotis Kolios provided supervision throughout.
	\end{itemize}

	\bibliographystyle{splncs04}
	\bibliography{bibliography}
	
\end{document}